%% file: main.tex
\documentclass[16pt,notitlepage,oneside]{report}

\usepackage{buetcseugthesis}
\begin{document}






\input{chapter_1}

\input{chapter_2.tex}

\input{chapter_3.tex}

\input{chapter_4.tex}

\input{chapter_5.tex}



\input{buetcseugthesisbibliography.tex}


\appendix
\input{appendix.tex}


\end{document}

%% file: chapter_1.tex
\chapter{Introduction}\label{chapter_1}

\section{Overview}

\begin{figure}[!t]
  \centering
  \includegraphics[width=0.9\textwidth]{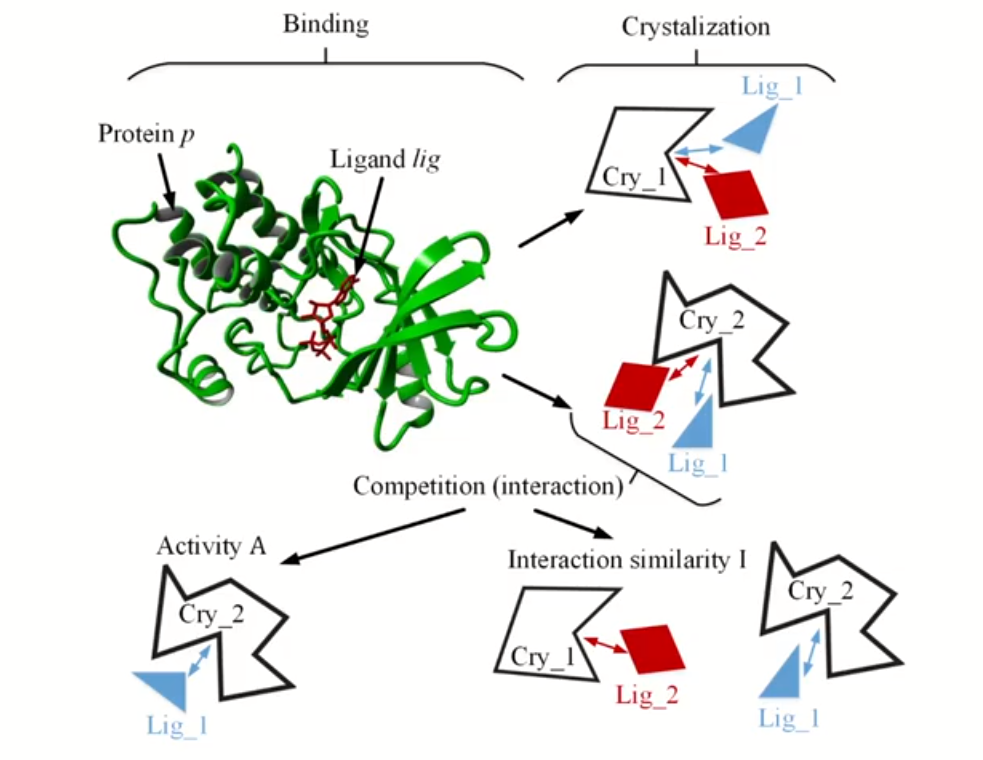}
  \caption{Visualization of Drug-target interaction}
  \label{visualization}
\end{figure}

The identification of novel drug–target (DT) interactions is a substantial part of the drug discovery process. Most of the computational methods that have been proposed to predict DT interactions have focused on binary classification, where the goal is to determine whether a DT pair interacts or not. This particularly does not help much towards determining whether a certain drug is able to inhibit certain target protein. If the drug is able to inhibit the target, then what quantity of the drug is needed to do that cannot be determined by this binary classification. So, a binary classification does not help answering all of these questions. On the other hand, protein–ligand interactions assume a continuum of binding strength values, also called binding affinity. This value is good indication of drug–target (DT) interactions and also denotes how much of the drug is needed to inhibit the target protein. But predicting this value still remains a challenge. \Cref{visualization} shows how a drug interacts with a target protein.\\

\section{Problem Statement}

\begin{figure}[!b]
  \centering
  \includegraphics[width=0.9\textwidth]{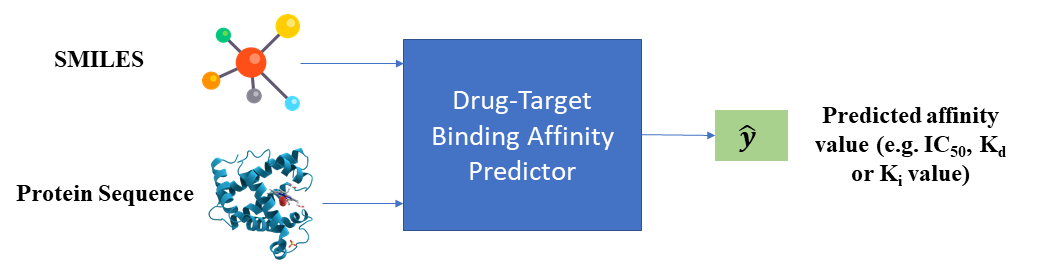}
  \caption{Binding Affinity Prediction Problem}
  \label{problem_statement}
\end{figure}

In this work we wanted to address the problem of predicting the drug-target binding affinity value using computation. Normally we intake a drug as a cure for certain diseases.
What these drugs actually do is that they bind with very specific or certain target proteins making it unable to function and  cause the diseases. Most of the cases a single drug often bind with multiple type of protein as often the diseases are caused by multiple protein rather than a single protein. Besides, each drug has some toxicity value and solubility value. These values are important to denote whether the drug is actually usable. If a drug can inhibit a target protein but requires a large amount of intake then probably, we do not use that drug for trial and other phases. \Cref{problem_statement} illustrate the problem statement nicely,\\

\textbf{SMILES~\cite{weininger1988smiles}:} SMILES means simplified-molecular input line entry system. SMILES is a specification in the form of a line notation for describing the structure of chemical species using short ASCII strings. SMILES strings can be imported by most molecule editors for conversion back into two-dimensional drawings or three-dimensional models of the molecules. The original SMILES specification was initiated in the 1980s. It has since been modified and extended. In 2007, an open standard called OpenSMILES\footnote{https://en.wikipedia.org/wiki/Open\_standard} was developed in the open-source chemistry community. In our work we used the SMILES string for the drug molecule. The dataset we used is called the KIBA dataset~\cite{KIBA} and it has a total of 2,111 drug SMILES string. These drugs have a total of 64 unique characters.\\

\textbf{Protein Sequence:} Protein sequencing is the practical process of determining the amino acid sequence of all or part of a protein or peptide. This may serve to identify the protein or characterize its post-translational modifications. Typically, partial sequencing of a protein provides sufficient information (one or more sequence tags) to identify it with reference to databases of protein sequences derived from the conceptual translation of genes. In this problem we use such protein sequence of humans that we want to inhibit using drugs. These proteins cause health issues when certain type of disease occurs and inhibiting them becomes necessary to get rid of the diseases. In the KIBA dataset there are total of 229 such human proteins are available. The protein sequences are in string format and has a total of 25 unique characters.\\

\textbf{Binding Affinity:} Binding affinity is the strength of the binding interaction between a single biomolecule (e.g. protein or DNA) to its ligand/binding partner (e.g. drug or inhibitor). There are numerous metrics that denotes the drug-target binding affinity value. 
\begin{itemize}
    \item \textbf{IC50:} Half-maximal inhibitory concentration (IC50) is the most widely used and informative measure of a drug's efficacy. It indicates how much drug is needed to inhibit a biological process by half, thus providing a measure of potency of an antagonist drug in pharmacological research.
    \item \textbf{$K_D$:} $K_D$ is determined experimentally and is a measure of the affinity of a drug for a receptor. More simply, the strength of the ligand–receptor interaction. To determine $K_D$, a fixed mass of membranes (with receptor) are incubated with increasing concentrations of a radioligand until saturation occurs.
    \item \textbf{$K_i$:} For noncompetitive inhibition of enzymes, the $K_i$ of a drug is essentially the same numerical value as the IC50, whereas for competitive and uncompetitive inhibition the $K_i$ is about one-half that of the IC50's numerical value.

\end{itemize}

Each of these metrics have some merits and demerits from one another. So, in the KIBA dataset they tried to take the merits of all these metrics and combine them into a KIBA score and the predictor’s task is to predict this score. The lower the KIBA score the better the drug is to inhibit the protein with little amount of intake.\\

\section{Motivation}

Drug discovery is the process through which potential new medicines are identified. It involves a wide range of scientific disciplines, including biology, chemistry and pharmacology. Historically, drugs were discovered by identifying the active ingredient from traditional remedies or by serendipitous discovery, as with penicillin. More recently, chemical libraries of synthetic small molecules, natural products or extracts were screened in intact cells or whole organisms to identify substances that had a desirable therapeutic effect in a process known as classical pharmacology. After sequencing of the human genome allowed rapid cloning and synthesis of large quantities of purified proteins, it has become common practice to use high throughput screening of large compounds libraries against isolated biological targets which are hypothesized to be disease-modifying in a process known as reverse pharmacology. Hits from these screens are then tested in cells and then in animals for efficacy.\\

Modern drug discovery involves the identification of screening hits, medicinal chemistry and optimization of those hits to increase the affinity, selectivity (to reduce the potential of side effects), efficacy/potency, metabolic stability (to increase the half-life), and oral bioavailability. Once a compound that fulfills all of these requirements has been identified, the process of drug development can continue. If successful, clinical trials are developed.\\

Modern drug discovery is thus usually a capital-intensive process that involves large investments by pharmaceutical industry corporations as well as national governments (who provide grants and loan guarantees). Despite advances in technology and understanding of biological systems, drug discovery is still a lengthy, expensive, difficult, and inefficient process" with low rate of new therapeutic discovery. It takes about 10 years of time and \$2.6B dollars to develop a single drug. \\

\begin{figure}[H]
  \centering
  \includegraphics[width=0.9\textwidth]{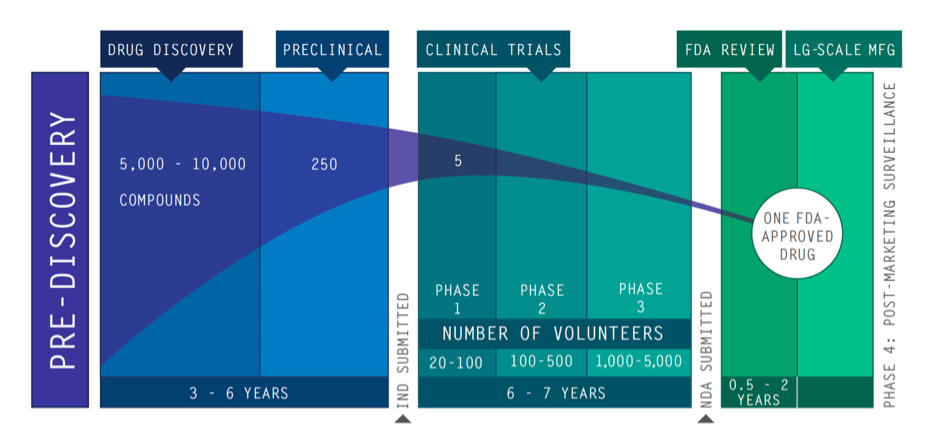}
  \caption{Drug Discovery Process in the USA}
  \label{drug_discovery_timeline}
\end{figure}

Currently we are going through a pandemic and we cannot wait ten years in a regular interval of lockdowns to have a drug. We need to find some way to discover drugs as quickly as possible. \\

Drug discovery is basically a search problem where we need to search a drug that cure a disease with little amount of intake. So, we can think that we are in a chemical maze where we need to find appropriate path to find the ideal drug. But doing that is not easy, there are $10^60$ ~\cite{kirkpatrick2004chemical} drug like molecule can be available and the whole USA industry can test about $10^5$ drugs per day. So, if we want to search the ideal drug for a disease in current setup it will be impossible if we want to brute force all the possible solution. So, we need to use computation in the drug discovery process.\\

The identification of novel drug–target (DT) interactions is a substantial part of the drug discovery process. Most of the computational methods that have been proposed to predict DT interactions have focused on binary classification, where the goal is to determine whether a DT pair interacts or not. However, protein–ligand interactions assume a continuum of binding strength values, also called binding affinity and predicting this value still remains a challenge. The increase in the affinity data available in DT knowledge-bases allows the use of advanced learning techniques such as deep learning architectures in the prediction of binding affinities. It will play an important role in the virtual screening for drug discovery.\\

\section{Limitation}
We know that the identification of drug-target binding affinity is a substantial part of the drug discovery process but it is not the only thing that solve the drug discovery pipeline and improve its timeline. There are several other factors that play a significant role in the drug discovery pipeline. For instance, there is Lipinski's rule of five, also known as Pfizer's rule of five or simply the rule of five (RO5), is a rule of thumb to evaluate drug likeness or determine if a chemical compound with a certain pharmacological or biological activity has chemical properties and physical properties that would make it a likely orally active drug in humans. The rule was formulated by Christopher A. Lipinski in 1997, based on the observation that most orally administered drugs are relatively small and moderately lipophilic molecules~\cite{lipinski1997experimental, lipinski2004lead}.\\

Lipinski analyzed all orally active FDA-approved drugs in the formulation of what is to be known as the Rule-of-Five or Lipinski's Rule. The Lipinski's Rule stated the following:

\begin{itemize}
    \item Molecular weight $< 500$ Dalton
    \item Octanol-water partition coefficient (LogP) $< 5$
    \item Hydrogen bond donors $< 5$
    \item Hydrogen bond acceptors $< 10$
\end{itemize}

Though our work predicts the binding affinity value as an end-to-end system but it does not consider these RO5 for a particular drug. So, for this the whole process is not completely automated at this moment. We can additionally address this constrain in future work of ours.\\

\section{Summary}

Predicting drug-target binding affinity is very problematic and time consuming for drug discovery process. Accurate and quick detection of binding affinity could increase the efficiency of this process. So, the predicting algorithms had to be computationally powerful and at the same time as precise as possible. The neural network approaches provided the desired level of computational power required for analyzing binding affinity. As for being precise, it was required to build the algorithm in such way that it would make the minimum number of mistakes. This is where the machine learning or deep learning algorithm came to help, as this model automatically traces back its own steps and adjusts itself to become more efficient and accurate. \\

%% file: chapter_2.tex
\chapter{Literature Review}\label{chapter_2}

\section{Overview}

In the recent years, there have been a lot of development in computational method to predict the drug-target binding affinity prediction. Most of these developments came from the increasing development in the data science field. The approaches that have been used for this work can be categorized in three sections. These are – 

\begin{itemize}
    \item Machine Learning based approaches
    \item Deep Learning based approaches using string representation
    \item Deep Learning based approaches using Molecular Graph
\end{itemize}

Each of these methods gives a lot of new ideas and ways to increase the evaluation results. But no methods are completely perfect as each one has some merits and demerits. In this section we will briefly know about the merits and demerits of these approaches and give some solutions to those demerits to overcome them.\\

\section{Machine Learning Based Approach}

\textbf{KronRLS~\cite{kronrls}:} It aims to minimize the following function, where $f$ is the prediction function

\begin{equation}
J(f)=\sum_{i=1}^{m}\left(y_{i}-f\left(x_{i}\right)\right)^{2}+\lambda\|f\|_{k}^{2}
\label{kronrls_equ}
\end{equation}

$\|f\|_{k}^{2}$ is the norm of $f$, which is related to the kernel function $k$, and $\lambda > 0 $ is a regularization hyper-parameter defined by the user. A minimizer for the equation can be defined as follows ~\cite{kimeldorf1971some}

\begin{equation}
f(x)=\sum_{i=1}^{m} a_{i} k\left(x, x_{i}\right)
\label{kronrls_equ_2}
\end{equation}

where $k$ is the kernel function. In order to represent compounds, they utilized a similarity matrix computed using Pubchem structure clustering server (Pubchem Sim\footnote{http://pubchem.ncbi.nlm.nih.gov} ), a tool that utilizes single linkage for cluster and uses 2D properties of the compounds to measure their similarity. As for proteins, the Smith–Waterman algorithm was used to construct a protein similarity matrix~\cite{smith1981identification}.\\

\textbf{SimBoost:} SimBoost is a gradient boosting machine based method that depends
on the features constructed from drugs, targets and drug–target pairs~\cite{simboost}. The proposed methodology uses feature engineering to build three types of features: (i) object-based features that utilize occurrence statistics and pairwise similarity information of drugs
and targets, (ii) network-based features such as neighbor statistics, network metrics (betweenness, closeness etc.), PageRank score, which are collected from the respective drug–drug and target–target networks (In a drug–drug network, drugs are represented as nodes and connected to each other if the similarity of these two drugs is above a user-defined threshold. The target–target network is constructed in a similar way.) and (iii) network-based features that are collected from a heterogeneous network (drug–target network) where a node can either be a drug or target and the drug nodes and target nodes are connected to each other via binding affinity value. In addition to the network metrics, neighbor statistics and PageRank scores, as well as latent vectors from matrix factorization
are also included in this type of network.\\

These features are fed into a supervised learning method named gradient boosting regression trees ~\cite{chen2016xgboost,chen2015higgs} derived from gradient boosting machine model~\cite{friedman2001greedy}. With gradient boosting regression trees, for a given drug–target pair $dt_i$, the binding affinity score $\overline{y_i}$ predicted as follows~\cite{simboost}:

\begin{equation}
\bar{y}_{l}=\theta\left(d t_{i}\right)=\sum_{m=1}^{M} f_{m}\left(d t_{i}\right), f_{m} \in F
\label{simboost_equ_1}
\end{equation}

in which $M$ denotes the number of regression trees and F represents the space of all possible trees. A regularized objective function to learn the set of trees $f_m$ is described in the following form~\cite{simboost}:

\begin{equation}
R(\theta)=\sum_{l} l\left(y_{i}, \bar{y}_{l}\right)+\sum_{m} \alpha\left(f_{m}\right)
\label{simboost_equ_2}
\end{equation}

where $l$ is the loss function that measures the difference between the actual binding affinity value $y_i$ and the predicted value $\overline{y_i}$  while $\alpha$ is the tuning parameter that controls the complexity of the model. The details are described in~\cite{chen2016xgboost,chen2015higgs, simboost}. Similar to ~\cite{kronrls}, ~\cite{simboost} also used PubChem clustering server for drug similarity and Smith–Waterman for protein similarity computation.\\

As we can see that both of these methods use hand engineering techniques along with machine learning techniques. Most of these methods score in a low coefficient correlation value which question the reliability of these methods. \newpage

\section{Deep-learning Based Approach Using String Representation}

\textbf{DeepDTA~\cite{deepdta}:} comprises two separate CNN blocks, each of which aims to learn representations from SMILES strings and protein sequences. For each CNN block, DeepDTA used three consecutive 1D-convolutional layers with increasing number of filters. The second layer had double and the third convolutional layer had triple the number of filters in the first one. The convolutional layers were then followed by the max-pooling layer. The final features of the max-pooling layers were concatenated and fed into three FC layers. DeepDTA used 1024 nodes in the first two FC layers, each followed by a dropout layer of rate 0.1. The third layer consisted of 512 nodes and was followed by the output layer.\\

\textbf{MT-DTI~\cite{mtdti}:} Molecule Transformer Drug Target Interaction (MT-DTI), based on a new molecule representation. They use a self-attention mechanism to learn the high-dimensional structure of a molecule from a given raw sequence. Their self-attention mechanism, Molecular Transformer (MT), is pre-trained on publicly available chemical compounds (PubChem database) to learn the complex structure of a molecule. This pre-training is important, because most datasets available for DTI training has only 2000 molecules, while the data for pre-training (PubChem database) contains 97 million of molecules. Although it does not contain interaction data but just molecules, their MT is able to learn a chemical structure from it, which will be effectively utilized when transferred to MT-DTI. Therefore, they transfer this trained molecule representation to their DTI model so that it can be fine-tuned with a DTI dataset.\\

\textbf{Deep-CPI~\cite{deepcpi}:} In their work they demonstrate that predicting compound-protein interaction (CPI) problem can be solved using a neural attention mechanism~\cite{bahdanau2014neural}. This mechanism allows them to consider which subsequences in a protein are important for a drug compound to predict CPIs (i.e. interaction sites) by using weights, which are also learned in the proposed neural networks. Furthermore, by using the obtained weights, the neural attention mechanism provides clear visualizations, which makes models easier to analyze even when modeling is performed using real-valued vector representations rather than discrete features. In their experiments using three CPI datasets ~\cite{liu2015improving, mysinger2012directory}, they demonstrated that the proposed approach based on end-to-end learning of GNN and CNN can achieve competitive or higher performance than existing approaches.\\

\textbf{WideDTA~\cite{widedta}:} They propose a methodology to predict protein-ligand binding affinity through text-only information of both proteins and compounds. Without relying on 3D structure information of the complex or 2D representation of the compound, they learn high dimensional features from sequences of the proteins and ligands. They suggested that, since the full-length sequence was used, the biologically important short subsequences that would be more powerful at representing the protein were lost due the
low signal to noise ratio~\cite{deepdta}. In order to overcome this problem, they propose to integrate different pieces of text-based information in the WideDTA model to provide a better representation of the interaction. They still utilize the protein sequence and ligand SMILES string by representing them as a set of words. A word of a protein sequence corresponds to a three-residue subsequence, whereas a word of a ligand is equal to an 8-character subsequence extracted with a sliding window approach~\cite{vidal2005lingo}. In addition, they use two textual information sources that can provide valuable clues about the specificity of the interaction.

The first piece of information we add to our words is protein motif and/or domain information. They utilize the PROSITE database~\cite{sigrist2010prosite} to extract motifs and profiles that are associated with a biologically significant function and domain. Then, they benefit from a recent study that showed that maximum common substructures (MCS) of ligands constitute the actual words in the chemical space~\cite{wozniak2018linguistic}. Approximately 100K MCS were used to extract a new set of words from the ligands. Together, these four text-based information sources constitute the WideDTA model.\\

\textbf{GANsDTA~\cite{gansdta}:} They propose a GANs~\cite{goodfellow2014generative}-based method to predict binding affinity, called GANsDTA for short. Their method comprises two types of networks, two partial GANs for the feature extraction from the raw protein sequences and SMILES strings separately and a regression network using convolutional neural networks for prediction. The contributions of their paper mainly include: They proposed a semi-supervised framework for DTA prediction; they adopted GAN to extract features of protein sequence and compound SMILES in an unsupervised way. Therefore, the proposed model can accommodate unlabeled data for the training as feature extractor using GANs does not require labeled data. This semi-supervised mechanism enables more datasets even without labels available for our model to learn proteins drugs features, leading to better feature representation and prediction performance accordingly. \\

\textbf{AttentionDTA~\cite{attentiondta}:} They propose a model to predict the binding affinities of drug-protein interactions with attention model using only sequences (1D representations) of proteins and drugs. After CNN extracted the abstract matrix representation of drugs and proteins, they use the attention module to calculate the scores between drugs and proteins representations at different positions, which allows them to consider which subsequences in a protein are more important for subsequences in the drug when predicting its affinity score and vice versa. They combine these representations to feed into a fully connected layer block, so called AttentionDTA. \\

All these methods use 1D convolutional neural network. One of the common problem these models face the loss of information in the initial layers. So, the representation in each stream remains under-represented as it can’t retain information when the input propagates through the model. In ~\cite{widedta} they try to address this problem by using additional data along with the input data.\\

\section{Deep-learning Based Approach Using Graph Representation}

These approaches use molecular graph as the input rather than the 1D string or the SMILES representation of the drug. Currently, these methods are becoming very much popular as they are able to create better representation for this task. \\

\textbf{DeepH-DTA~\cite{deephdta}:} They introduce a novel deep-learning-based DTI framework, called deepH-DTA, which geometrically exploits existing the topological structure of drug molecules as input features, along with the corresponding molecular fingerprints. They investigated twofold prediction of chemical interactions between protein sequences of target and homogeneity of drug candidate compounds. To this end, they propose a novel DTI prediction framework that utilizes a HGAT ~\cite{wang2019heterogeneous} for efficient modeling of interactions of various targeted topological representation of drugs. Simultaneously, they introduce two layers of bidirectional ConvLSTM~\cite{shi2015convolutional} to capture spatio-sequential characteristics of drug sequences encoded in a simplified molecular-input line-entry system (SMILES) format~\cite{weininger1988smiles}. The spatio-sequential sequences capture both positional features of input SMILES and the long-term dependency representation within input sequences.\\

\textbf{GraphDTA~\cite{nguyen2021graphdta}:} GraphDTA is a new neural network architecture capable of directly modelling drugs as molecular graphs. The approach is based on the solution they submitted to the IDG-DREAM Drug-Kinase Binding Prediction Challenge\footnote{https://www.synapse.org/\#!Synapse:syn15667962/wiki/583305}, where they were among the Top Ten Performers from 530 registered participants\footnote{https://www.synapse.org/\#!Synapse:syn15667962/wiki/592145}. Their results suggest that graph neural networks are highly accurate, abstract meaningful concepts, and yet fail in predictable ways. They conclude with a discussion about how these insights can feedback into the research cycle.\\

\textbf{GIN~\cite{wang2020dipeptide}:} They propose a novel feature extraction method which is polypeptide frequency of word frequency based on natural language word frequency characteristics to enhance the ability of protein sequence expression. The network model is constructed by merging the graph convolutional network that calculates the graph structure of drugs and the convolutional neural network that calculates the hidden relationship of protein features. The results of output are combined as the input of two hidden layers for regression training and prediction of DTA.\\

\textbf{DeepGS~\cite{lin2020deepgs}:} Their work is the first to consider both local chemical context and topological structure to learn the interaction between drugs and targets. Their experimental results demonstrate that considering jointly local chemical context and topological structure are effective and the newly designed molecular structure modeling approach works well under their proposed framework.\\

While these works provide much greater results than the rest of the works but to understand and comprehend their work and in some cases to use their work the knowledge in molecular science and chemical domain is required.\\

\section{Summary}

In this chapter, we have looked into all sorts of work that has been done in the drug target binding affinity prediction domain. We have looked into a lot of works that has been done in the last ten years and some of the works achieve great results but they do possess some demerits as well. We have briefly talked about those demerits in this chapter and we will try to address those in the following chapter with our contribution in this research topic.

%% file: chapter_3.tex
\chapter{Methodology}\label{chapter_3}

\section{Overview}

Studying the related works helped me to decide which approach should be taken. All the necessary information was gathered form these works that help me to formulate the design the architecture of the model and to implement it in test cases. From background study, we learned about Simplified molecular-input line-entry system (SMILES) and Protein Sequence as well. We also learned about the Neural Network model and various methods to deep learning and machine learning. These numerous methods that we went through, allowed us to shape a methodology of our own.

\begin{figure}[H]
  \centering
  \includegraphics[width=0.9\textwidth]{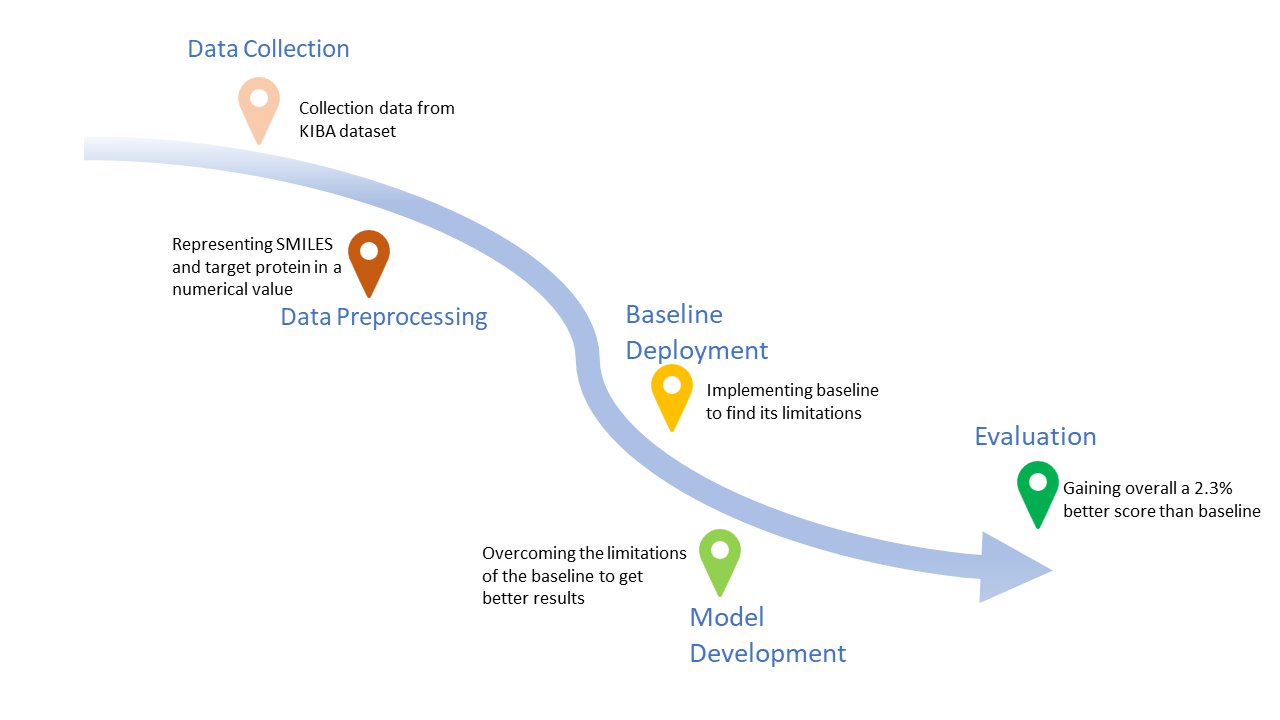}
  \caption{Methodology Development Road-map}
  \label{methodology}
\end{figure}

\Cref{methodology} shows the major milestones of this project. Throughout this chapter we are going to take a systematic and analytical look at each step of this project, discussing the principals and theories behind our applied methods. We will give an overview of our model and show which methods we used in each layer.\\

\section{Contribution}

In the baseline work of DeepDTA~\cite{deepdta} we saw that they used a two-stream network one for getting the drug representation and one for getting protein representation. Then finally concatenating these two representations together to the fully connected layers. But drug lengths and the sequence lengths aren’t same. Most of the cases protein sequences are much larger than the drugs SMILES length. Therefore, the model is not capable to fully comprehend the information in the forward layers after concatenation layer. Therefore, in our work we introduce another stream where we took the final representation of the drug and protein before global max-pooling and then concatenated their representation and pass them in a separate stream with more filters. Then finally concatenating the three streams representation before the fully connected layers.\\

Beside using a separate stream, we also used residual connection in each stream taking motivation from ResNet architecture~\cite{he2016deep}. It helped making the representation of each stream more robust so that it can hold the information in the fully connected layers and make compensation for the information loss that was occurring in the baseline. These application in our network helped overcoming the limitation of the baseline and improved the results on the test data.\\

\section{Dataset}

We evaluated our proposed model on KIBA dataset~\cite{KIBA}, which were previously used as benchmark datasets for binding affinity prediction evaluation.\\

The KIBA dataset, originated from an approach called KIBA, in which kinase inhibitor bioactivities from different sources such as $K_i$, $K_D$ and $IC50$ were combined. KIBA scores were constructed to optimize the consistency between $K_i$, $K_D$ and $IC50$ by utilizing the statistical information they contained. The KIBA dataset originally comprised 467 targets and 52,498 drugs. In ~\cite{simboost} the dataset was filtered it to contain only drugs and targets with at least 10 interactions yielding a total of 229 unique proteins and 2,111 unique drugs. \Cref{dataset_summary} summarizes this dataset in the forms that we used in our experiments

\begin{table}[H]
  \begin{center}
    \caption{Summary of the Dataset}
    \label{dataset_summary}

    \begin{tabular}{|c|c|c|c|}
      \hline
      \textbf{Dataset} & \textbf{Protein} & \textbf{Compound} & \textbf{Interaction} \\
      \hline
      KIBA & $229$ & $2,111$ & $1,18,254$ \\
      \hline
    \end{tabular}
  \end{center}
\end{table}

\begin{figure}[H]
  \centering
  \includegraphics[width=0.9\textwidth]{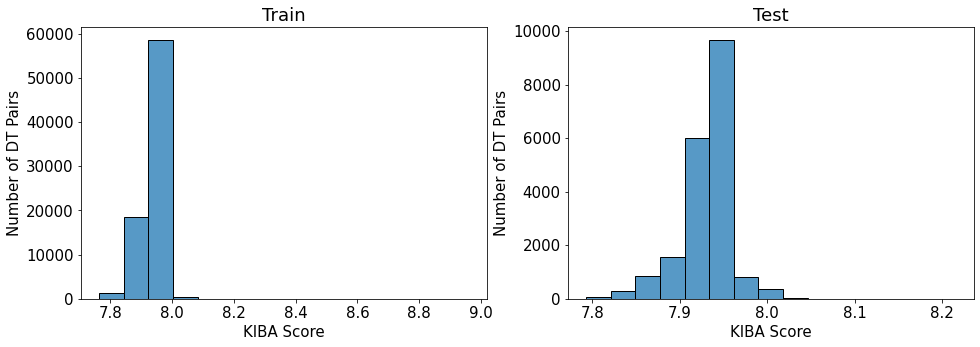}
  \caption{Distribution of KIBA Score}
  \label{kiba_score_dist}
\end{figure}

The distribution of the KIBA scores is depicted in \Cref{kiba_score_dist} for the training and test set. In ~\cite{simboost} KIBA scores was pre-processed as follows: 

\begin{itemize}
    \item For each KIBA score, its negative was taken
    \item The minimum value among the negatives was chosen
    \item The absolute value of the minimum was added to all negative scores
\end{itemize}

Thus, constructing the final form of the KIBA scores.\\

\begin{figure}[!b]
  \centering
  \includegraphics[width=0.9\textwidth]{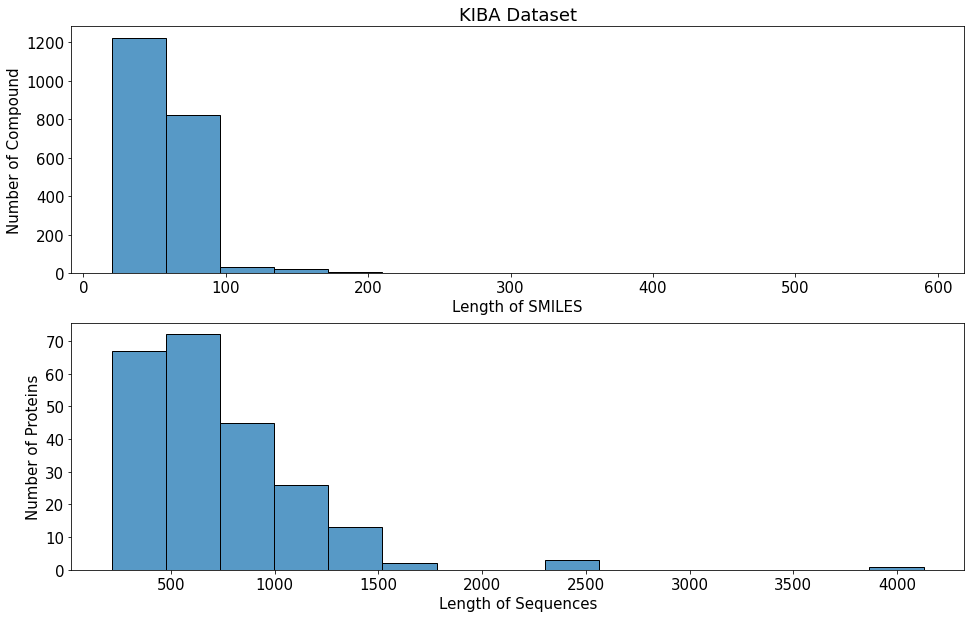}
  \caption{Summary of the KIBA Dataset}
  \label{kiba_summary}
\end{figure}

The compound SMILES strings of the KIBA dataset were extracted from the Pubchem compound database based on their Pubchem CIDs ~\cite{bolton2008pubchem}. For, first the CHEMBL IDs were converted into Pubchem CIDs and then, the corresponding CIDs were used to extract the SMILES strings. \Cref{kiba_summary} illustrates the distribution of the lengths of the SMILES strings of the compounds in the KIBA datasets. For the compounds of the KIBA dataset, the maximum length of a SMILES is $590$, while the average length is equal to $58$.\\

The protein sequences of the KIBA dataset were extracted from the UniProt protein database based on gene names/RefSeq accession numbers~\cite{apweiler2004uniprot}. \Cref{kiba_summary} shows the lengths of the sequences of the proteins in the KIBA dataset. The maximum length of a protein sequence is $4,128$ and the average length is $728$ characters. \\

We should also note that the Smith–Waterman (S–W) similarity among proteins of the KIBA dataset is at most $60\%$ for $99\%$ of the protein pairs. This statistic indicate that the dataset is non-redundant.\\

\section{Data Pre-processing}

We used integer/label encoding that uses integers for the categories to represent inputs. We scanned approximately 2M SMILES sequences that we collected from Pubchem and compiled 64 labels (unique letters). For protein sequences, we scanned 550K protein sequences from UniProt and extracted 25 categories (unique letters).\\

Here we represent each label with a corresponding integer (e.g. ‘C’: 1, ‘H’: 2, ‘N’: 3 etc.). The label encoding for the example SMILES, ‘CN=C=O’, is given below.\\

\begin{center}
\textbf{[ C\hspace{2em}N\hspace{2em}=\hspace{2em}C\hspace{2em}=\hspace{2em}O]\hspace{2em}$=$\hspace{2em}[1\hspace{2em}3\hspace{2em}63\hspace{2em}1\hspace{2em}63\hspace{2em}5]}\\
\end{center}

Protein sequences are encoded in a similar way using label encodings. Both SMILES and protein sequences have varying lengths. Hence, in order to create an effective representation form, we decided on fixed maximum lengths of 100 for SMILES and 1000 for protein sequences for the dataset. We chose these maximum lengths based on the distributions illustrated in \Cref{kiba_summary} so that the maximum lengths cover at least 80\% of the proteins and 90\% of the compounds in the dataset. The sequences that are longer than the maximum length is truncated, whereas shorter sequences are zero-padded. 

\section{Proposed Architecture}

To solve the problem of determining the binding affinity value using the SMILES string and the Protein Sequence string, we used a deep neural network using three streams and convolutional neural network block along with residual skip connection. In this section we will discuss the whole proposed model architecture and their different section and the hyperparameters used for the model.\\

\begin{figure}[!tb]
  \centering
  \includegraphics[width=1\textwidth]{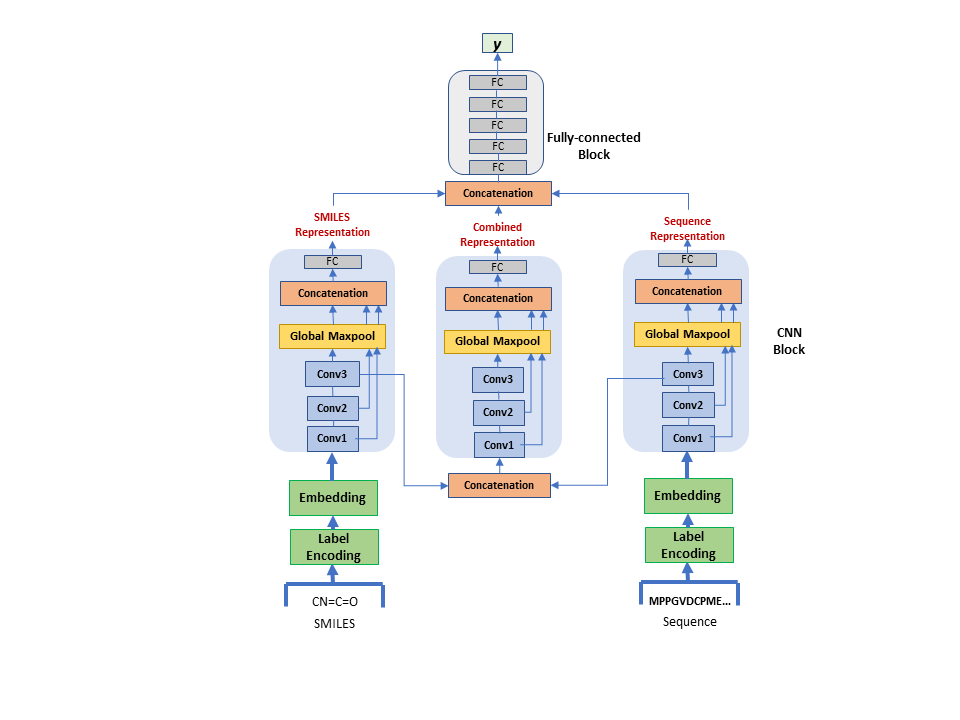}
  \caption{Proposed Model Architecture ResDTA}
  \label{model_architecture}
\end{figure}

The proposed model architecture is a multi-stream network where first stream is used to create the representation for the drug, second streams create a combined representation and the final stream creates the representation for the protein sequence. Then these three representations are concatenated and used as an input to the fully-connected block. After the fully connected block we get the affinity value as output. Since we used residual skip connection in the CNN block therefore, we named our model as ResDTA. \Cref{model_architecture} shows the whole architecture of our proposed model.\\

Now, we will discuss about the various blocks used in the proposed model with all the hyperparameters used in different layers. Since the proposed model is a multi-stream network therefore, first I will discuss about the SMILES stream and the Protein Sequence stream cause they both contain the same blocks and later I will discuss about the combined stream,\\

\textbf{Label Encoding:} in label encoding the data was pre-processed. Deep neural network cannot work with string type data, so we represented the string in the way described in the data pre-processing section so that it can be used in the following layers. After the label encoding layers, the input will have a tensor of size 100 integer for SMILES stream and 1000 integer for sequence stream. So, for SMILES stream we get the following output from the label encoding,

\begin{equation}
l\left(X^{S M I L E}\right)=\left\{x \mid x \in N^{100}\right\}
\end{equation}

Where $l(.)$ represents the label encoder,  $X^SMILE$ represents the SMILES string and $N^100$ represents a tensor of integers of size one hundred. Similarly for the sequence stream, 

\begin{equation}
l\left(X^{Sequence}\right)=\left\{x \mid x \in N^{1000}\right\}
\end{equation}

\textbf{Embedding Layer:} It is a simple lookup table that stores embeddings of a fixed dictionary and size. This module is often used to store word embeddings and retrieve them using indices. The input to the module is a list of indices, and the output is the corresponding word embeddings. This layer takes a tensor of integer as input and using its dictionary it converts that into one-hot encoding tensor then pass that tensor in a linear transformation layer used in a fully connected block. For SMILES stream we use a dictionary of size 64 as there are total of 64 unique character that can be used to represent SMILES and for similar reason the dictionary size was 25 for sequence stream. For linear transformation the output was a tensor for 128. The SMILES and sequence embedding representation is shown in the following two equation accordingly,\\


\begin{equation}
\operatorname{Em}\left(l^{n \times 100}\right)=\left\{x \mid x \in R^{n \times 128 \times 100}\right\}
\end{equation}

\begin{equation}
\operatorname{Em}\left(l^{n \times 1000}\right)=\left\{x \mid x \in R^{n \times 128 \times 1000}\right\}
\end{equation}

Where, $Em(.)$ represents embedding layer function and  $l^{n \times i}$ represents the output of the label encoding and n represents the batch size in each mini-batch.\\

\textbf{CNN Block:} CNN represents convolutional neural network. In this block we have used 1D convolutional over a signal composed of 128 signal planes. Each 1D convolution is followed by a ReLU activation and also a Global Maxpool. For the 1D convolution in the simplest case, the output value of the layer with input size $(N,C_{in} ,L)$ and output $(N,C_{out} ,L_{out})$ can be precisely described as:

\begin{equation}
\operatorname{out}\left(n_{i}, C_{\text {out } j}\right)=\operatorname{bias}\left(C_{\text {out } j}\right)+\sum_{k=0}^{C_{\text {in }}-1} \text { weight }\left(C_{\text {out } j}, k\right) \star \operatorname{input}\left(n_{i}, \mathrm{k}\right)
\end{equation}

where $\star$  is the valid cross-correlation operator, $n$ is a batch size, $C$ denotes a number of channels, $L$ is a length of signal sequence. In each CNN block we have used three sequential 1D convolution where we passed the Embedding output. We have used 32 filters in the first convolution then used 64 filters and 96 filters for the next two 1D convolution accordingly, as for padding we use no padding. We used a stride of 1 and kernel size of 8 for the filters in both stream\\

\begin{center}
   $Input: (n, C_{in}, L_{in}) or (C_{in}, L_{in})$\\
   $Output: (n, C_{out}, L_{out}) or (C_{out}, L_{out})$
\end{center}

\begin{equation}
   L_{out}=\lfloor\frac{L_{in }+2 \times{ padding }-{ dilation } \times({ kernel\_size }-1)-1}{ stride }+1\rfloor 
\end{equation}

Here, $C$ represents the channels of the convolution and $L$ represents the features of the channel.\\

The activation we used after convolution is ReLU. Each convolution is followed Global Maxpool which downsamples the input representation by taking the maximum value over the channel dimension. To apply this global maxpool we used skip connection or residual connection so that while having the final representation the information in the former layers is retained. At the end of the three convolution and the global maxpool all the features are concatenated together and passed through a linear transformation layer to get the final representation from the both streams of the same size.\\


\begin{equation}
\mathrm{C}_{\text {SMILES }}\left(E^{n \times 128 \times 100}\right)=\left\{x \mid x \in R^{n \times 256}\right\}
\end{equation}

\begin{equation}
\operatorname{C_{Sequence}}\left(E^{n \times 128 \times 1000}\right)=\left\{x \mid x \in R^{n \times 256}\right\}
\end{equation}

Here, $C(.)$ represents the CNN block and $E^{n\times i \times j}$ represents output of the embedding layer.\\

After CNN block in the two streams, we get the SMILES representation and the sequence representation. Now before going into details about the fully-connceted block let us first look at the combined stream and discuss about how do we get the combined representation. In this stream we begin with the output of the last convolution from the other two streams. Then we concatenate the features so that we get a tensor of the size such as $X \in R^{n \times 96 \times 1058}$  . Then we pass this tensor to a CNN block. This CNN has the similar component as the other streams but the only difference is the number of filters used by the three convolution blocks. It has 192, 288 and 96 filters for the three convolutional layer accordingly. The number of filters for the convolution layers were determined by looking at the results of the cross-validation of different folds. As this contains information about both the drug and also the target sequence therefore the output of this stream is twice as big as the other two stream. So, the output of this stream is such that $X \in R^{n \times 512}$.

Finally, we get the output of each stream (SMILES stream, combined stream and the sequence stream) and concatenate them before using it as an input to the fully-connected block. Therefore, we get a tensor of 1024 dimension as the input of the fully-connected block. The fully-connected block consist of five linear transformation layers with various output features. The transformation function is as follows,\\

\begin{equation}
    y = xA^T + b
\end{equation}

Here $y$ is the output feature and $x$ is the input feature, $A$ is the trainable weight matrix and $b$ is bias. For the five fully connected layer we used the out features as 2048, 2048, 1024, 512 and 1 accordingly. The number of layers and their output features were selected by observing the cross-validation results of different folds. All the layers have an activation of ReLU after the linear transformation but the last layer. Last layer outputs the binding affinity value therefore no activation is used in this layer. During training to reduce the bias we attached a Dropout with every fully-connected layer. Dropout is used during training to randomly zeroes some of the elements of the input tensor with probability $p$ using samples from a Bernoulli distribution. Each channel will be zeroed out independently on every forward call. Furthermore, the outputs are scaled by a factor of $\frac{1}{1-p}$ during training. This means that during evaluation the module simply computes an identity function. In our case the value $p$ was kept as $0.1$ for all the cases where Dropout was used.\\


\begin{equation}
\hat{K}=F_c\left(C_{S M I L E S}\left(E^{n \times 128 \times 100}\right) \oplus C_{\text {Sequence }}\left(E^{n \times 128 \times 1000}\right) \oplus C_{\text {Combined }}\left(\left\{X \mid X \in R^{n \times 96 \times 1058}\right\}\right)\right)
\end{equation}

Here, $\widehat{K}$ represents the predicted binding affinity value and in our work, it is the predicted KIBA score, $F_c (.)$ represents the fully-connected block and $\oplus$ is used to represent concatenation.

%% file: chapter_4.tex
\chapter{Result and Discussion}\label{chapter_4}

\section{Overview}

In any Machine Learning or Deep Learning application it is important to evaluate the network in order to understand how the network will work when it comes across real world data or future data. This is the main purpose for any predictive modeling work. In order to evaluate any model, we need to select the evaluation metric first. The choice of evaluation metric is always based on the task the model is doing. For instance, for a task like classification problem a common evaluation metric will be accuracy whereas for a regression task a common evaluation metric would be mean square error. In our task we used different evaluation metric to verify our model generalization and performance gain over the baseline models which will be discussed at later section of this chapter. In addition, we will also discuss about the training hyper-parameters and training setup of our experiments so that anyone can reproduce our work.\\

\section{Evaluation Metrics}

As we discussed earlier evaluation metrics are a key factor to understand if a predictive model is working correctly as the way they should and will it work similarly for future data. It is always a good practice not to rely on a single metric for evaluating a model cause no metric is completely perfect and cannot interpret the model performance perfectly. Our task is a regression task so here we cannot use accuracy as there are no class involved in this task. We chose different evaluation metrics for evaluating the performance of our model. These metrics are also well suited for our work and also have been used in the previous works. It helped us evaluate the model further with the baseline methods. The computation of these metrics is as follows.

\textbf{Concordance Index (CI)~\cite{gonen2005concordance}:} This metric is computed using the following equation,

\begin{equation}
C I=\frac{1}{Z} \sum_{\delta_{i}>\delta_{j}} h\left(b_{i}-b_{j}\right)
\end{equation}

where $b_i$ is the prediction value for the larger affinity $\delta_i$, $b_j$ is the prediction value for the smaller affinity $\delta_j$, $Z$ is a normalization constant, $h(x)$ is the step function described as the following equation,\\

\begin{equation}
h(x)=\left\{\begin{array}{cc}
1, & \text { if } x>0 \\
0.5, & \text { if } x=0 \\
0, & \text { if } x<0
\end{array}\right.
\end{equation}

The metric measures whether the predicted binding affinity values of two random drug–target pairs were predicted in the same order as their true values were. We used paired-t test for the statistical significance tests with 95\% confidence interval (the larger value of CI indicates better model performance). \\

\textbf{Mean Squared Error (MSE):} MSE is a common evaluation metric used for regression task. It represents the average of differences between predicted and actual output values. So, the smaller value of MSE a model generates the better that model is performing is assumed. MSE is calculated using the following equation\\

\begin{equation}
MSE=\frac{1}{n} \sum_{i=1}^{n}\left(P_{i}-Y_{i}\right)^{2}
\end{equation}

Here, $P$ is the prediction vector, and $Y$ corresponds to the vector of actual outputs. $n$ indicates the number of samples.\\

\textbf{R squared $r_m^2$ :} This denotes the external prediction performance of the model. Meanwhile, the model is acceptable only when $r_m^2 > 0.5$ and,\\

\begin{equation}
r_{m}^{2}=r^{2} \times\left(1-\sqrt{r^{2}-r_{o}^{2}}\right)
\end{equation}

here $r^{2}$ and  $r_{o}^{2}$ designate the squared correlation coefficient parameters for the predicted and actual values with and without intercept.\\

\section{Model Training}

Model training is a crucial step for any predictive models to learn properly. Without training properly, we cannot expect the model to give appropriate prediction in the future or unseen data. Besides, to reproduce the results of a model we need to know specifically how the models were trained before testing it. Below we will describe and discuss about the different hyper-parameter we used during our training and various training technique that we used for achieving better results from the baseline-\\

\textbf{Optimizer:} To ensure the model learns the best way possible from the data, some hyper-parameters needed to be specified correctly. In the case of the neural network, we need to consider non-convex optimizers as a neural network that can have more than one local optima. It is very hard to determine which of these optima is the global optima. So, while we were training our ResDTA model, we needed to focus on finding the global optima from the loss surface of the network. To ensure a global optimum we needed to keep in mind a few things. such as the value of learning rate, not getting stuck on local optima, the changing morphology of Neural Network loss surface. Considering these two-point we used the Adam~\cite{kingma2014adam} optimizer. Adaptive moment estimation optimizer utilizes two algorithms: root means squared prop and Gradient descent with momentum. It allows the optimizer to produce smooth gradients. It combines the parameters from the two methods with two of its hyper-parameters:  learning rate and epsilon. The starting learning rate was set to 0.0001 and the epsilon value was 1e-08.\\

\textbf{Loss function:} Loss or the cost function is the main function that determines how model weights should be updated and using the loss value optimizer updates the weights of the model using gradient descent. So, determining the appropriate loss function for the model is a very important task. A popular loss function for classification task is categorical cross-entropy. But since our work is a regression task so we used root mean square error as our loss function to train our model. Previously we discuss about the mean-square error and how that is important for the model that does regression task. But our incentive to use RMSE is we trained our model for many epochs and with higher epoch training loss tend to go very low whereas there is a lot learning has to be done. But as the loss become low the training and the weights update gets slowed as well. We were facing this issue using the MSE loss but as we used RMSE it fixed the problem as the loss does not go very low at early stage of the training and as an overall outcome, we got better weights for our model at later stage of the training also. \\

\begin{equation}
R M S E=\sqrt{\frac{1}{n} \sum_{i=1}^{n}\left(P_{i}-Y_{i}\right)^{2}}
\end{equation}

\textbf{Learning rate scheduler:} As the training progress the training has to slowed to reach to the global optimal in the loss plane. Therefore, the learning rate needs to be reduced when the loss is reaching towards the global optima. The learning rate was reduced ten-fold after every 200 epochs of training. Empirically we found that after every 100 epochs of training reinitializing the optimizer and loading the best model weights and resuming the training from there gave better results over training the model continuously for couple of epochs at a stretch.\\

\textbf{Gradient Accumulation:} Gradient descent works best if we supply all of the available training data in a single forward pass. But this can’t be done for the most part due to the computational limitation. Therefore, we usually divide our data in mini-batches and go through them one by one. The network predicts batch labels, which are used to compute the loss with respect to the actual targets. Next, we perform backward pass to compute gradients and update model weights in the direction of those gradients. Gradient accumulation modifies the last step of the training process. Instead of updating the network weights on every batch, we can save gradient values, proceed to the next batch and add up the new gradients. The weight update is then done only after several batches have been processed by the model. Thus, gradient accumulation helps to imitate a larger batch size on a less expensive machine. Empirically we saw that gradient accumulation always helps the model during moments when training seems to stuck at a certain position. It improves the model training and therefore improve in the model performance as well.\\

\begin{figure}[H]
  \centering
  \includegraphics[width=0.9\textwidth]{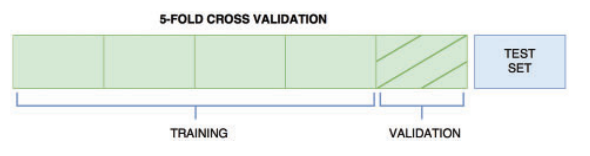}
  \caption{Train-validation-test setup}
  \label{train_setup}
\end{figure}

\textbf{Hyper-parameters and other training specification:} The remaining hyper-parameters that are used during training are the number of epochs and the mini-batch size. We trained our model for a total of 400 epochs among them 200 epochs were trained with a learning rate of 0.0001 and later 200 epochs were trained using a learning rate of 0.00001. Throughout the training the mini-batch size was always kept at 256. In order to learn a generalized model, we randomly divided our dataset into six equal parts in which one part is selected as the independent test set. The remaining parts of the dataset were used to determine the hyper-parameters via 5-fold cross validation. \Cref{train_setup} illustrates the partitioning of the dataset. The same train and test folds were used by the other baselines as well for a fair comparison.\\

For writing the codes we used the popular deep-learning framework Pytorch\footnote{https://pytorch.org/}  and for training we used Kaggle\footnote{https://www.kaggle.com/}  kernels with a GPU.\\

\section{Model Evaluation}

In this section first, we will be looking into the various model we tried out before we landed on ResDTA. We used the Concordance Index (CI) to evaluate between these models. Then at the later part of this section we will be looking how ResDTA performed in comparison with the baseline models using the 1D string representation of the drug and the target protein sequence. Finally, we will be looking into how ResDTA’s prediction stands against the measured or actual binding-affinity value for the dataset.\\

\begin{figure}[H]
  \centering
  \includegraphics[width=0.9\textwidth]{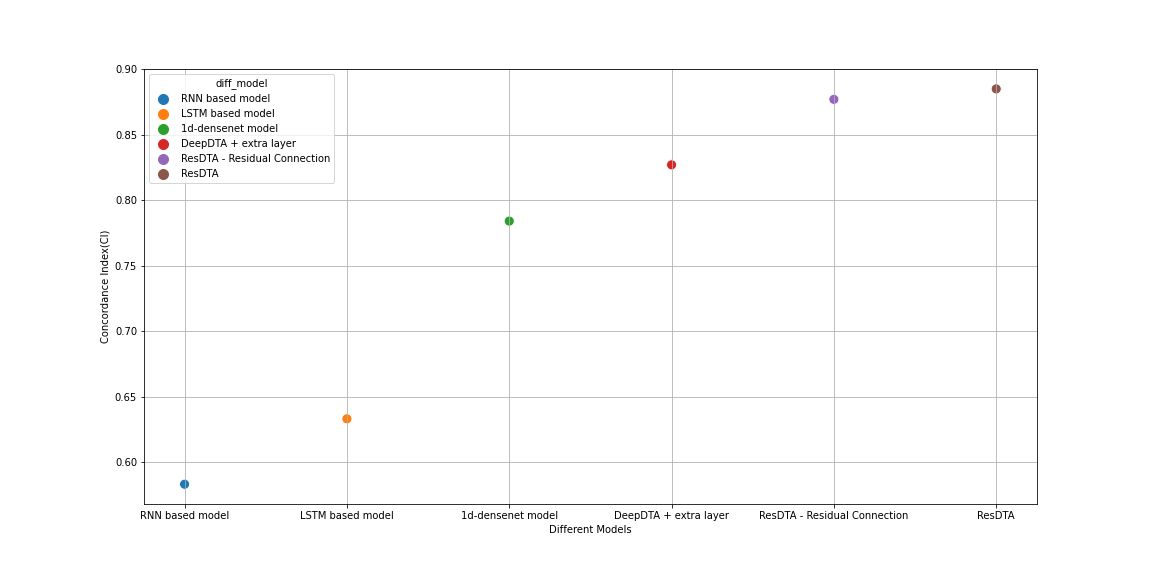}
  \caption{Model Evaluation}
  \label{model_eval}
\end{figure}

Since the input data was a string of sequence both for the drug and the target and traditionally recurrent neural networks (RNN) work better on sequence data. So, we begin our experiment with RNN at the very beginning. But we got a very low CI score for the model. We tried out tuning a lot of hyper-parameters but the model gave prediction very much randomly therefore we could not improve the score. Then we tried out long-short time memory (LSTM) networks~\cite{lstm}. LSTM is proven to perform the best among all the recurrent type of model in task that contains time-series data. Though our work did not contain time-series data, we tried LSTM for our work. As a result, we did not find that the model was working properly as we got very low CI score. Then we move from recurrent networks and starts using convolutional neural networks specifically we used the 1d convolutional neural network. In the computer vision task, various 2d convolutional neural network have been in use. As our data can be processed only by 1d convolution so we converted one of the popular vision models 2d convolutional model DenseNet~\cite{huang2017densely} into 1D DenseNet. 1d DenseNet did perform a lot better than the recurrent networks but it was still way behind the baseline CI score. So, next we used the baseline DeepDTA model with some extra fully-connected layers and increase the number of filters in each stream. But it did not improve the reported result. Then we used the baseline model and added combined stream to the model and finally we get a performance gain in CI score. And finally we landed on the ResDTA network by attaching residual skip connection with the combined stream into the model which gave a significant performance gain from the baseline. \Cref{model_eval} illustrates the results of the different models of our experiments\\

For demonstrating the competitiveness of our model, we conduct an end-to-end comparison with the cutting-edge approaches (either machine- or deep-learning approaches) adopted for predicting affinity scores, and we conducted the comparative experiments under the same conditions for fair comparison with our method.\\

In \Cref{model_eval_table}, we provide the average obtained Concordance Index (CI) score, Mean Square Error (MSE) score and rm2 score to each study on the datasets, respectively. It can be noted that machine-learning models such as KronRLS~\cite{kronrls} and SimBoost~\cite{simboost} show worse performance compared to other deep-learning approaches. This is owing to their dependence on similarity matrices between drugs and targets as well as hand-crafted features. On the other hand, deep-learning techniques that automatically capture feature representation show great performance improvement.\\

\begin{table}[!b]
  \begin{center}
    \caption{Model comparison with cutting edge approaches on the KIBA dataset} 
    \label{model_eval_table}

    \begin{tabular}{|c|c|c|c|}
      \hline
      \textbf{Models} & \textbf{CI(std)} & \textbf{MSE} & \textbf{$r_{m}^{2}(std)$} \\
      \hline
      \multicolumn{4}{|c|}{\textbf{Machine Learning Based Approaches}} \\
      \hline
      KronRLS~\cite{kronrls} & 0.782 (0.001) & 0.411 & 0.342 (0.001) \\
      \hline
      Simboost~\cite{simboost} & 0.836 (0.001) & 0.222 & 0.629 (0.007) \\
      \hline
      \multicolumn{4}{|c|}{\textbf{Deep Learning Based Approaches}} \\
      \hline
      DeepDTA~\cite{deepdta} & 0.863 (0.002) & 0.194 & 0.673 (0.009) \\
      \hline
      MT-DTI~\cite{mtdti} & 0.882 (0.001) & 0.220 & 0.584 (0.003) \\
      \hline
      DeepCPI~\cite{deepcpi} & 0.852 (0.002) & 0.211 & 0.657 (0.004) \\
      \hline
      WideDTA~\cite{widedta} & 0.875 (0.001) & 0.179 & 0.675 (0.005) \\
      \hline
      GANsDTA~\cite{gansdta} & 0.866 (0.001) & 0.224 & 0.775 (0.008) \\
      \hline
      Attention-DTA~\cite{attentiondta} & 0.882 (0.004) & 0.162 & 0.735 (0.003) \\
      \hline
      \hline
      ResDTA(w/o Residual Connection) & 0.877 (0.001) & 0.0002 & 0.653 (0.005) \\
      \hline
      \textbf{ResDTA} & \textbf{0.885 (0.001)} & \textbf{0.0002} & 0.671 (0.004) \\
      \hline
    \end{tabular}
  \end{center}
\end{table}

Among the deep learning based approaches, Attentio-DTA~\cite{attentiondta} and MT-DTI~\cite{mtdti} yielded best results with CI of 0.882, MSE of 0.220 and 0.162 respectively. It can be observed that the proposed ResDTA has a robust performance on the datasets, achieving 0.885 (0.003 improvement) for CI and 0.0002 (reduced by 0.1618) for MSE and achieved 0.671 in $r_m^2$ and we discussed that a model will be accepted when that it has a score of  $r_m^2 > 0.5$ , so ResDTA will be accepted based on that statement. It means the prediction generated by ResDTA is not by fluke and it has actual statistical significance. The result indicates the superiority of our proposed approached compared to the most recent studies for predicting DTA. Accordingly, we observe that our model outperforms existing deep-learning methods on two measures, which can be explained due to several factors: 

\begin{itemize}
    \item In comparing with DeepDTA~\cite{deepdta} we show that our model is able to retain the information in the final representation of SMILES and sequence using the residual connection. 
    \item In comparing with WideDTA~\cite{widedta} we show that using combined stream gives much better performance than using additional information along with the data.
    \item In comparing with Attention-DTA~\cite{attentiondta} we show that residual connection gives much better performance than using attention.
\end{itemize}

\begin{figure}[!b]
  \centering
  \includegraphics[width=0.9\textwidth]{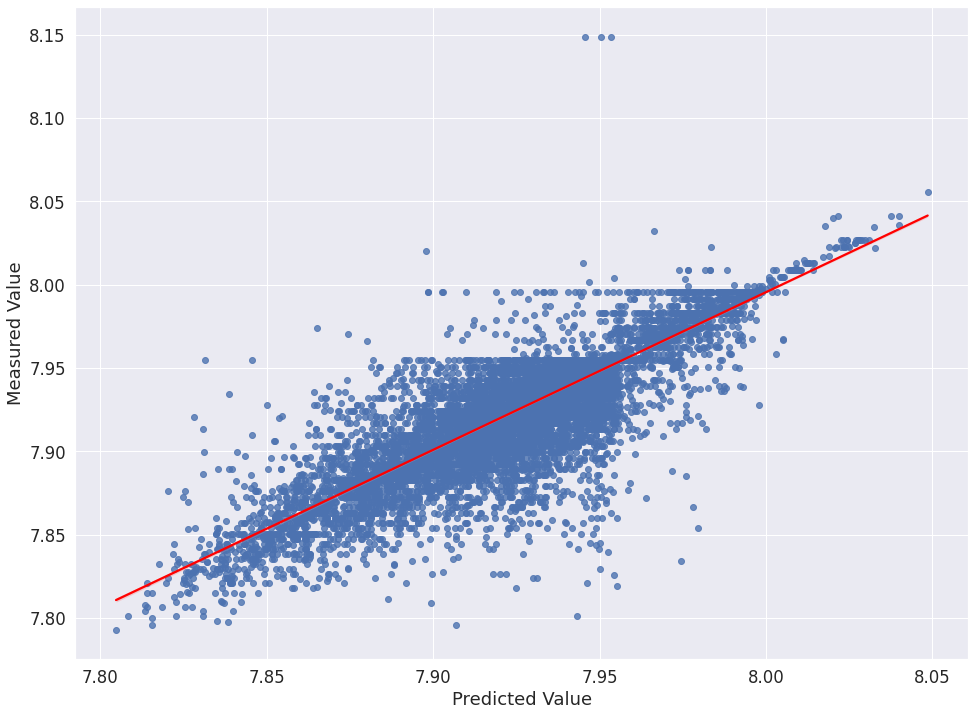}
  \caption{Predictions from our model against measured (real) binding affinity values for the KIBA dataset (KIBA score).}
  \label{reg_plot}
\end{figure}

Generally, the obtained results and comparisons demonstrate that our model achieves competitive performance outperforms against these baselines methods. Moreover, \Cref{reg_plot} present the scatter plots of the proposed model predicted affinity score against the actual measured value on the datasets. The model achieves better performance when the estimated affinity scores are close to the original scores, and hence the instances should appear close to the red line. Principally, for the datasets, the data instances are close to the red regression line which, in turn, demonstrates that the proposed architecture has a competitive prediction performance.\\

\section{Summary}

The proposed model was able to predict the drug-target binding affinity value much better than all the baseline that we have considered in this work. The overall concordance index score was 0.887 and mean square error was 0.0002 which is better than all the baseline. This performance was achieved on a previously unseen dataset. So, it is evident that the model was properly trained in predicting affinity value.

%% file: chapter_5.tex
\chapter{Conclusion and Future Work}\label{chapter_5}

We propose a deep-learning based approach to predict drug–target binding affinity using only sequences of proteins and drugs. We use Convolutional Neural Networks (CNN) along with the residual skip connections (ResDTA) to learn representations from the raw sequence data of proteins and drugs and also use a combined stream in our network to build an overall robust representation and finally used fully connected layers in the affinity prediction task. We compare the performance of the proposed model with recent studies that used 1D string representation of the protein and drugs. We experimented on the KIBA dataset~\cite{KIBA}.\\

Our experiments show that using residual skip connections works much better than attention mechanism. It creates robust representation in each of the stream of the network. We also show that using the combined stream taking input from the last convolution layer from each of the stream creates a combined representation which reduce the necessity to use additional information along with the input data. The results show that the prediction that our model is producing has some statistical significance and it is not generating prediction by accident or by fluke. Our result also shows an improvement from the current state-of-the-art work AttentionDTA~\cite{attentiondta} using 1D string representation for protein and drugs. (From CI score 0.882 to CI score 0.885).\\

The major contribution of this study is the presentation of a novel deep learning-based model for drug–target affinity prediction that uses only character representations of proteins and drugs. By simply using raw sequence information for both drugs and targets, we were able to achieve better performance than the baseline methods.\\

In future to further improve our result we can use natural language processing (NLP) models for protein embeddings (e.g. ProteinBERT~\cite{brandes2022proteinbert}). These models help to represent protein sequence in an appropriate embedding which is better than any other methods and also these embeddings are able to create much better representation if we passed it in our model’s protein stream. Besides, we can use generative model like autoencoders to generate the molecular graph from the SMILES string and use those graphs along with graph convolutional network to create a robust representation of the drug. Therefore, we would not need any additional domain knowledge for using molecular graph. So, incorporating all these along with our model we can definitely improve our result further in our future work.\\

A large percentage of proteins remains untargeted, either due to bias in the drug discovery field for a select group of proteins or due to their undruggability, and this untapped pool of proteins has gained interest with protein deorphanizing efforts~\cite{edwards2011too, friedman2001greedy, fedorov2010targeted, o2016ligand}. The methodology can then be extended to predict the affinity of known compounds/targets to novel targets/drugs as well as to the prediction of the affinity of novel drug–target pairs.\\

%% file: buetcseugthesisbibliography.tex

\clearpage
\renewcommand\bibname{References}
\addcontentsline{toc}{chapter}{References}

\bibliographystyle{ieeetr} 
\bibliography{buetcseugthesis}

%% file: appendix.tex
\chapter{Technical Details}\label{appendix_a}

Here we provide some technical details of the implementation. The project is being developed in Python 3.6.10 environment and makes use of the Kaggle\footnote{www.kaggle.com} Kernels with GPU and the following libraries ,platforms and hardware:

\begin{itemize}
    \item Keras v2.3.1, Deep Learning Library for Theano and Tensorflow
    \item Pytorch 1.11, Machine Learning Framework build on Python
    \item cv2 v4.5, a Python wrapper for OpenCV
    \item SciPy v1.5, a Python-based ecosystem of open-source software for mathematics, science and engineering
    \item NumPy v1.18.5, the fundamental package for scientific computing with Python
    \item Pandas v1.0.5, a data analysis and manipulation library for Python
    \item Matplotlib v3.2.2, a Python plotting library
    \item Seaborn v0.11.2, a statistical data visualization library for Python
    \item CUDA v10.0.5, a GPU computing platform
    \item CuDNN c10.0.5, the NVIDIA GPU-accelerated library
    \item Tesla P100-PCIE, a server GPU used for model training and provided by Kaggle
\end{itemize}